\documentclass[letterpaper, table]{article} 
\usepackage{aaai2026}
\usepackage{times}  
\usepackage{helvet}  
\usepackage{courier}  
\usepackage[hyphens]{url}  
\usepackage{graphicx} 
\usepackage{amsmath}
\usepackage{amssymb} 
\usepackage{booktabs} 
\usepackage{multirow}
\usepackage{threeparttable}
\usepackage{natbib}  
\usepackage{caption} 
\usepackage{algorithm}
\usepackage{algorithmic}

\usepackage{newfloat}
\usepackage{listings}
\DeclareCaptionStyle{ruled}{labelfont=normalfont,labelsep=colon,strut=off} 
\floatstyle{ruled}
\newfloat{listing}{tb}{lst}{}
\floatname{listing}{Listing}
\title{DyC-STG: Dynamic Causal Spatio-Temporal Graph Network for Real-time Data Credibility Analysis in IoT}
\author{
    Guanjie Cheng\textsuperscript{\rm 1}, Boyi Li\textsuperscript{\rm 2*}, Peihan Wu\textsuperscript{\rm 1}, Feiyi Chen\textsuperscript{\rm 1}, Xinkui Zhao\textsuperscript{\rm 1*}, Mengying Zhu\textsuperscript{\rm 1}, Shuiguang Deng\textsuperscript{\rm 1}
}
\affiliations{
    \textsuperscript{\rm 1}School of Computer Science and Engineering, Zhejiang University\\
    \textsuperscript{\rm 2}School of Computer Science and Engineering, Northeastern University\\

}

\usepackage{bibentry}

\begin{document}

\maketitle

\begin{abstract}
The wide spreading of Internet of Things (IoT) sensors generates vast spatio-temporal data streams, but ensuring data credibility is a critical yet unsolved challenge for applications like smart homes. While spatio-temporal graph (STG) models are a leading paradigm for such data, they often fall short in dynamic, human-centric environments due to two fundamental limitations: (1) their reliance on static graph topologies, which fail to capture physical, event-driven dynamics, and (2) their tendency to confuse spurious correlations with true causality, undermining robustness in human-centric environments. To address these gaps, we propose the Dynamic Causal Spatio-Temporal Graph Network (DyC-STG), a novel framework designed for real-time data credibility analysis in IoT. Our framework features two synergistic contributions: an event-driven dynamic graph module that adapts the graph topology in real-time to reflect physical state changes, and a causal reasoning module to distill causally-aware representations by strictly enforcing temporal precedence. To facilitate the research in this domain we release two new real-world datasets. Comprehensive experiments show that DyC-STG establishes a new state-of-the-art, outperforming the strongest baselines by 1.4 percentage points and achieving an F1-Score of up to 0.930.
\end{abstract}


\section{Introduction}

The rise of large models and agents is driving the Internet of Things (IoT) towards a new era of advanced autonomous intelligence, which in turn places unprecedented demands on the quality of its underlying data \cite{guo2024large,aouedi2024survey}. In the smart home scenario, this deep integration of Artificial Intelligence (AI) and IoT is transforming living spaces into vast and complex sensing networks \cite{huda2024experts}. The high-dimensional, multivariate time-series data streams continuously generated by these networks form the very backbone of advanced automated services, such as intelligent energy management \cite{nikpour2025intelligent}, proactive security \cite{rehman2024proactive}, and personalized scene recommendations \cite{xiao2023know}.

However, the performance and reliability of all such intelligent services hinge entirely upon the credibility of the data that drives them. Consequently, the issue of data credibility—that is, whether the data accurately reflects the true state of the physical world—has become a core bottleneck impeding the advancement of intelligent systems toward greater autonomy and reliability. While Spatio-Temporal Graph Neural Networks (STGNNs) have achieved tremendous success in modeling such dependencies, particularly in structured data scenarios like traffic flow forecasting \cite{ju2024cool,kong2024spatio}, their direct application to the smart home context reveals two fundamental limitations stemming from the unique dynamic and event-driven nature of the environment:
\begin{figure}[t]
    \centering
    \includegraphics[width=0.9\linewidth]{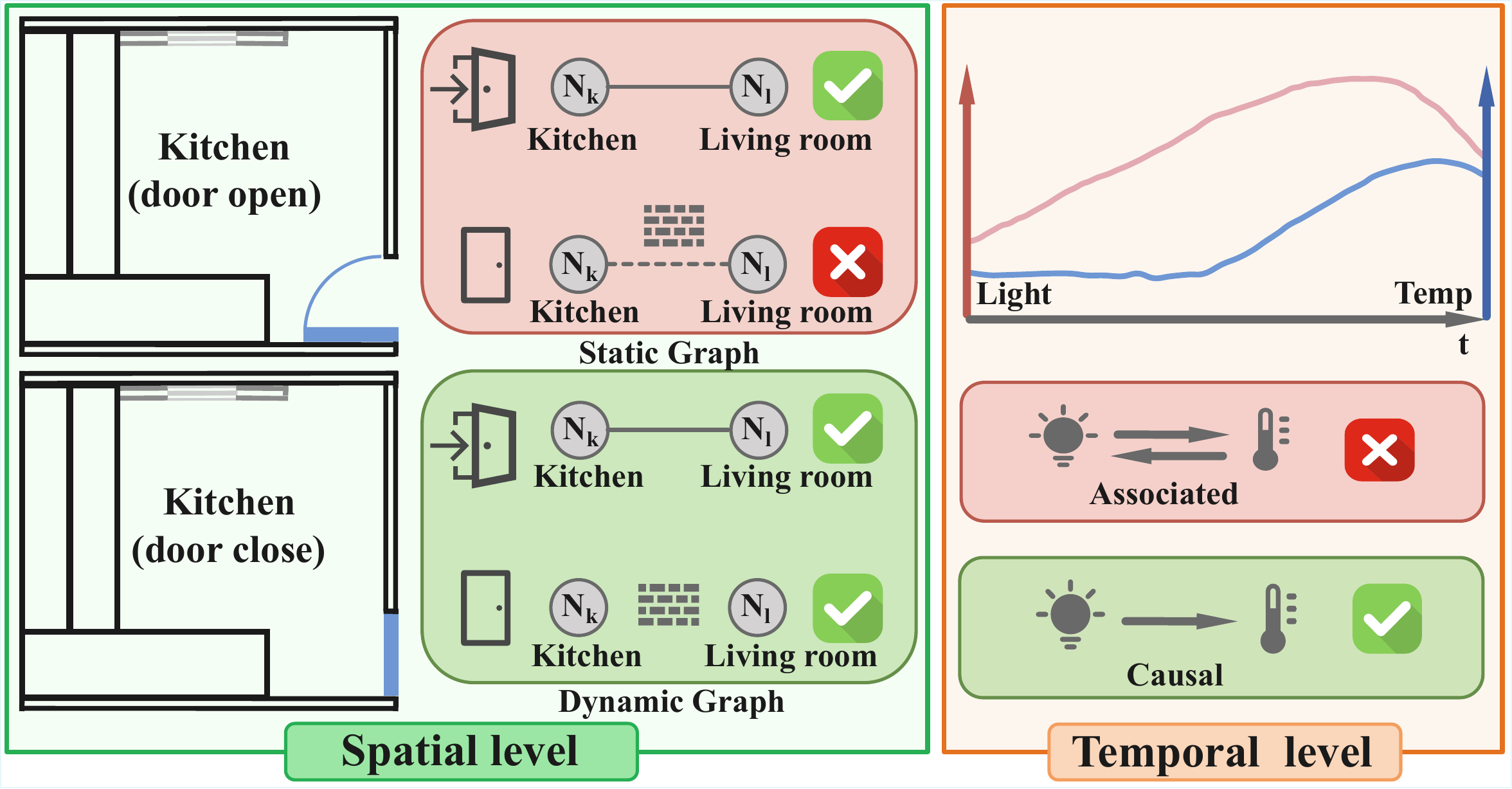}
    \caption{Illustration of the core challenges.}
    \label{fig1_introduction}
\end{figure}\\
1. \textbf{Static Assumption of Spatial Dependencies:} The majority of existing methods rely on a pre-defined, static graph topology to capture spatial correlations \cite{tang2023gadbench,pazho2023survey}. In a smart home, however, the physical relationships between sensors are dynamic. For instance, opening or closing a window fundamentally alters the correlation strength between indoor and outdoor temperature sensors. Even models that employ adaptive adjacency matrices \cite{chen2023multi,sun2025scalable} often learn graph structures that evolve too slowly to capture the abrupt topological changes triggered by discrete events. This assumption of a static or slowly-changing graph fundamentally limits the model's capacity to represent real-world physical dynamics.\\
2. \textbf{Causal Confusion in Temporal Dependencies:} 
In the complex environment of a smart home, a single human activity can trigger simultaneous changes across multiple sensors, creating inter-dependencies between variables in multivariate time series. While existing deep learning models are good at capturing these co-occurrence patterns, they often struggle to distinguish true causal chains from spurious correlations \cite{gong2024causal}. This ``correlation-causation confusion" is a critical flaw; a model that learns a spurious link between, for instance, a coffee machine and a toaster (which are often used together in the morning) may fail when only one is used. This disregard for the underlying causal mechanisms compromises the model's robustness, making it liable to misinterpret valid but less common sequences of events as anomalies.

Recent efforts have begun to tackle these issues from separate angles. To capture dynamic spatial dependencies, existing research \cite{geng2024stgaformer,liu2024spatial} have moved beyond static graphs by learning topologies that evolve based on data correlations. On the temporal front, another line of work \cite{fu2024causal,gong2024causal} has focused on explicit causal discovery, attempting to first learn a causal graph from data and then use it as a prior to guide the model's predictions. However, the dynamic graph models learn abstract correlations that are frequently ungrounded from the discrete, physical events that truly govern the environment. Meanwhile, approaches based on explicit causal discovery often rely on strong statistical assumptions and decouple the causal learning from the end-to-end representation learning process. A critical gap thus persists: the lack of a unified framework that marries physically-grounded, event-driven dynamic graphs with robust causal reasoning embedded within the architecture.

To address these challenges and bridge this gap, this paper proposes a novel framework named the Dynamic Causal Spatio-Temporal Graph Network (DyC-STG), designed as an end-to-end solution with deep reasoning capabilities for assessing data credibility in smart homes. Specifically, we design an event-driven dynamic graph construction module that dynamically reconstructs the graph topology based on the state of control nodes, thereby precisely capturing dynamic spatial dependencies. Furthermore, we introduce a causality-enhanced Transformer module \cite{wang2024spatiotemporal,akyurek2024context,nichani2024transformers} that fundamentally redefines the mechanism's temporal receptive field. Unlike a global, bidirectional field that captures all correlations, our approach imposes a causal structure where the receptive field for each time step is strictly confined to its historical context. This forces the model to learn directional, cause-and-effect relationships rather than mere temporal co-occurrence.

The main contributions are summarized as follows:
\begin{itemize}
\item We propose DyC-STG, a novel, physically-grounded framework for time-series analysis in dynamic IoT environments. It pioneers a new modeling paradigm by simultaneously capturing event-driven physical dynamics via a dynamic graph topology, and distilling robust, causally-aware representations through a causality-enhanced Transformer.
\item We design and propose a novel event-driven dynamic graph paradigm, explicitly reconstructing the graph topology in real-time by modulating edge weights based on the physical state of control nodes, thereby grounding the model's spatial reasoning in the physical world.
\item We introduce a novel causal reasoning paradigm that redefines the temporal receptive field of self-attention. By confining each time step's receptive field strictly to its historical context, it compels the model to distinguish directional cause-and-effect relationships from mere temporal co-occurrence.
\item To facilitate the research in this domain, we release two large-scale, real-world datasets (a 5 GB contribution). Extensive experiments on them demonstrate that DyC-STG establishes a new state-of-the-art in data credibility assessment. It surpasses the strongest baselines by a significant margin, achieving an F1-Score of 0.9297 and an AUC of 0.9886, which represent absolute improvements of 1.44 and 0.51 percentage points respectively, highlighting its superior performance and potential to address the ``trust crisis" in IoT data.
\end{itemize}

\section{Related Works}
Ensuring the credibility of massive Internet of Things (IoT) data is fundamental for advanced applications \cite{paramesha2024big}. Early statistical and machine learning methods for data quality assessment \cite{tang2023gadbench,pazho2023survey} are often inadequate for dynamic environments because they fail to model complex spatio-temporal dependencies.
\subsubsection{Spatio-Temporal Graph Models for IoT Data.}
Spatio-Temporal Graph Neural Networks (STGNNs) have rapidly evolved to capture these dependencies \cite{jin2023spatio}. First-generation models like DCRNN \cite{li2017diffusion}, STGCN \cite{yu2017spatio}, STFGNN \cite{li2021spatial}, and STGNCDE \cite{choi2022graph} fused graph convolutions with sequence models but were limited by static graph structures, a bottleneck for adapting to physical events in smart homes \cite{xie2023spatio,kong2024spatio}. Second-generation models such as GWNet, MTGNN, and AGCRN \cite{tian2021spatial,wu2020connecting,zheng2023spatio} introduced adaptive graphs, yet these learned topologies remain fixed post-training, lacking real-time responsiveness.Third-generation models shifted to attention and Transformers, with works like ST-MambaSync \cite{shao2025st}, QA-STGACN \cite{qiu2024integrating}, STFT \cite{wang2024spatiotemporal}, SSL-STMFormer \cite{li2025ssl}, and PDFormer \cite{jiang2023pdformer} dynamically weighting spatio-temporal dependencies. Their core limitation, however, is a tendency to capture all correlations, failing to distinguish true causal drivers from spurious associations. This ``correlation-causation confusion" critically undermines robustness in aperiodic, human-driven scenarios.

\begin{figure*}[t]
    \centering
    \includegraphics[width=0.95\linewidth]{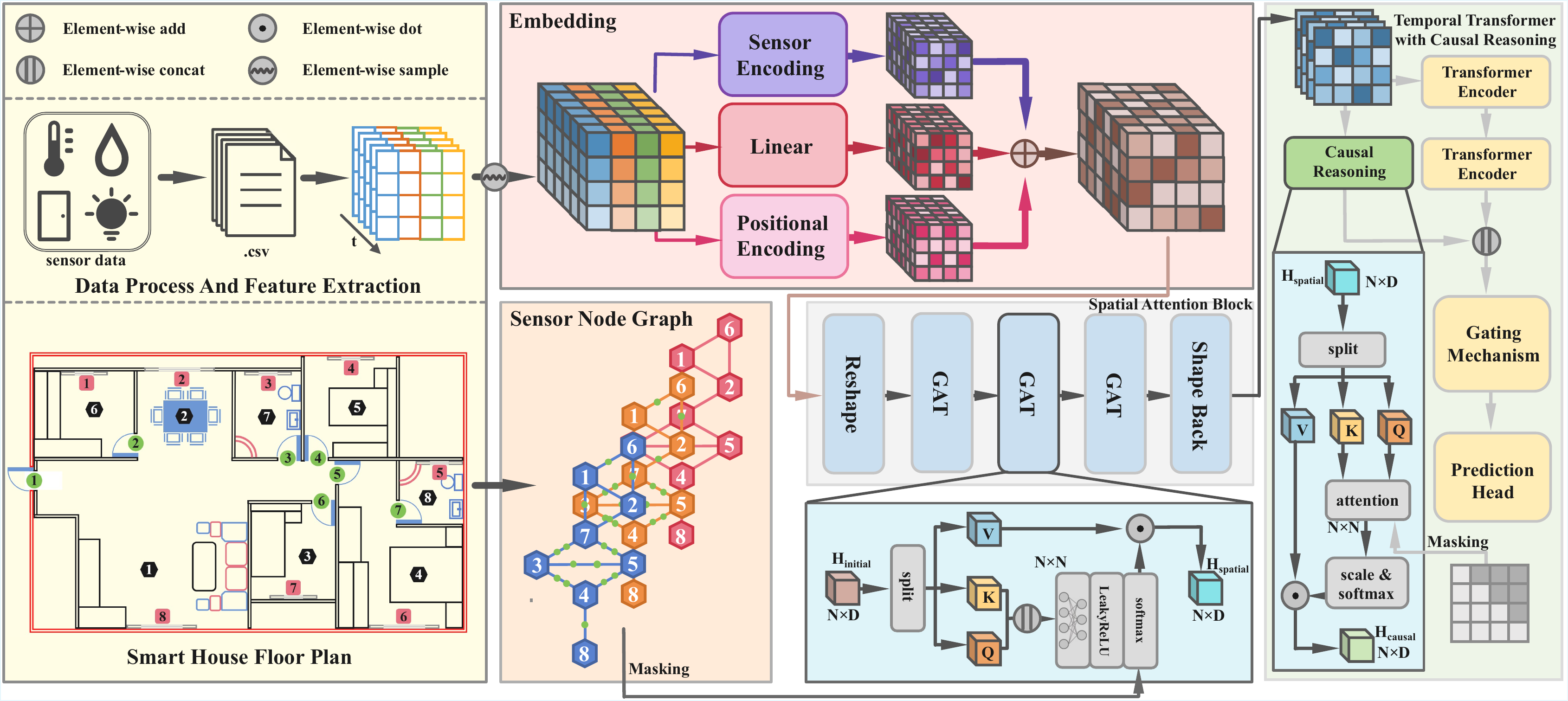}
    \caption{The Overall Architecture of the DyC-STG. The architecture adopts a cascaded design that decouples spatial and temporal modeling.  First, spatial information is aggregated independently at each time step via a stack of GAT layers, which operate on a dynamic graph structure derived from the physical environment.  Subsequently, the resulting spatially-aware features are fed into parallel branches for deep temporal modeling and causal reasoning.  Finally, a gating mechanism intelligently fuses these two representations before a prediction head generates the final output.}
    \label{fig2_introduction}
\end{figure*}

\subsubsection{Causal Reasoning in Time-Series Analysis.}
Distinguishing causality from correlation is central to model robustness in time-series analysis. Traditional methods like Granger causality tests are hampered by linearity and stationarity assumptions \cite{gong2024causal,kapoor2024latent}. While some deep learning works learn explicit causal graphs \cite{jing2024causality,zhang2024graph} or perform complex counterfactual reasoning \cite{wang2024counterfactual,verma2024counterfactual}, they are often too computationally prohibitive for efficient inference. A more pragmatic approach is enforcing causality through architectural design, such as using masking mechanisms to prevent information leakage from the future, thereby respecting the temporal arrow of causality \cite{chen2024using}.

Our DyC-STG framework is engineered to bridge these gaps. It integrates an event-driven dynamic graph for real-time physical adaptation with a causal Transformer that uses strict masking to enforce temporal causality. This novel fusion of Dynamic Physical Perception and Causal Logic Distillation offers a new paradigm to address the data credibility challenge in complex IoT systems.
\section{Methodology}
This section details the technical architecture of our proposed DyC-STG. We begin by formalizing the problem and then elaborate on the four core modules of our framework which is shown in Figure 2: (1) Dynamic Graph Construction; (2) Spatial Dependency Modeling; (3) Temporal Feature Extraction and Causal Refinement; (4) the Gated Fusion and Output Layer.
\subsection{Problem Formulation}
The primary objective of this work is to assess the credibility of data streams originating from a dynamically evolving smart home sensor network. We formally model the network's topological structure at any given time step as a time-varying graph $\mathcal{G}_t = (\mathcal{V}, \mathcal{E}_t)$, where $\mathcal{V}$ is the set of ${N}$ sensor nodes (i.e., $|\mathcal{V}| = {N}$), and $\mathcal{E}_t$ is the time-varying edge set representing their connectivity.\\
\indent Formally, the model has two primary inputs: (1) a multivariate time series tensor ${X}\in{R}^{B \times T \times N \times D_{\text{in}}}$, where ${B}$ is the batch size, ${T}$ is the length of the historical window, and ${D}_{in}$ is the dimension of raw and engineered features for each sensor at each time step; and (2) a sequence of adjacency matrices $A = \{A^1, A^2, \dots, A^T\}$ that represents the evolving graph topology.\\
\indent The goal is to learn a mapping function $f$ that takes the historical spatio-temporal features and the dynamic graph structure to predict the credibility of each data point:
\begin{equation}
    f \colon (X, A) \to Y
\end{equation}
The model outputs a credibility tensor $Y \in {R}^{B \times T \times N \times 1}$, where each element $Y_{b,t,n}$ represents the predicted probability that a sensor reading is trustworthy. This score is converted to a binary decision $D_{b,t,n}$ by applying a threshold $\zeta \in (0,1)$:
\begin{equation}
    D_{b,t,n} = 
    \begin{cases} 
        \text{Trustworthy} & \text{if } Y_{b,t,n} > \zeta \\
        \text{Untrustworthy} & \text{if } Y_{b,t,n} \le \zeta 
    \end{cases}
\end{equation}
To ensure robustness against class imbalance, the threshold $\zeta$ is not fixed but is empirically calibrated by maximizing the F1-Score on the validation set.

The final decision $D$ is a critical trigger for system-level actions. A determination of ``Untrustworthy" can initiate data quarantining and imputation, generate alerts for human verification, or signal the need for proactive hardware maintenance. This mechanism ensures that high-level autonomous services are shielded from corrupted data, thus preserving the integrity and reliability of the entire IoT ecosystem.

\begin{figure}[t]
    \centering
    \includegraphics[width=0.8\linewidth]{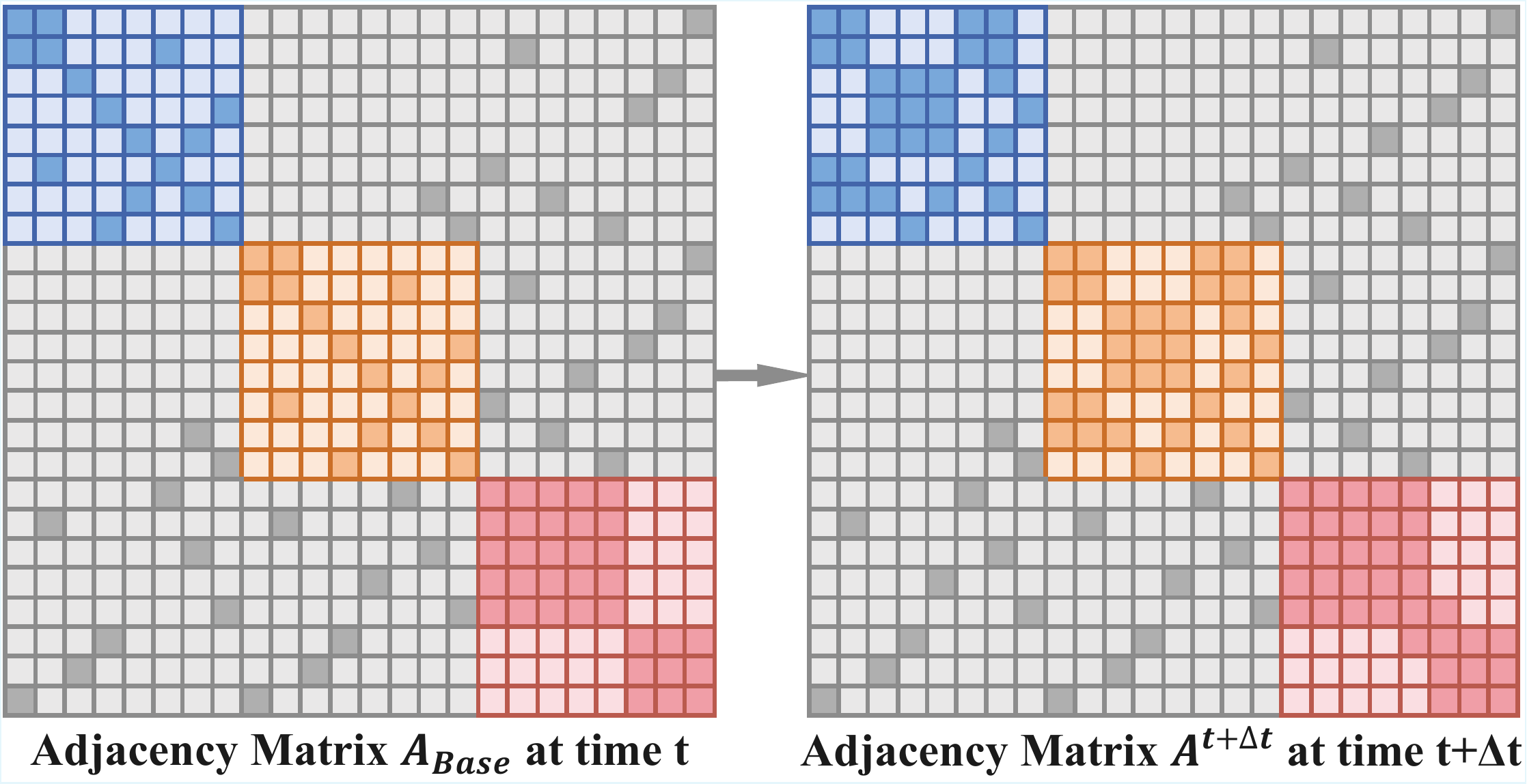}
    \caption{Illustration of the Dynamic Graph Construction mechanism. The process begins with a static base graph, $A_{\text{Base}}$ (left), which represents the latent physical topology. Upon a state change in a control node (e.g., a door opening, $s_c^{t+\Delta t}=1$), a modulation function activates latent pathways to produce the effective adjacency matrix $A^{t+\Delta t}$ (right). In this visualization, colored blocks represent dense intra-group connections (blue: temperature, orange: humidity, red: light), while gray blocks represent inter-group connections, which are dynamically activated.}
    \label{fig:dynamic_graph}
\end{figure}

\subsection{Dynamic Graph Construction}
A critical limitation of conventional spatio-temporal models is the assumption of a static graph, which fails to capture how discrete physical events alter the nature of sensor correlations. Our work posits that spatial dependencies are not static but are, in fact, conditional upon the dynamic state of the physical environment.

To encode this physical prior into our model, we introduce the Dynamic Graph Construction module as shown in Figure 3. This module generates a time-varying topology by modulating a latent static graph based on real-time environmental state changes, unfolding in a two-stage process for each time step $t$:

\subsubsection{1. Latent Static Topology.}
We first define a base adjacency matrix, $A_{\text{Base}} \in {R}^{N \times N}$. This matrix represents the \textit{latent}, time-invariant topology of the network, encoding all potential interaction pathways based on physical proximity or functional coupling.

\subsubsection{2. State-Driven Structural Modulation.}
We identify a subset of sensors as structural control nodes $\mathcal{V}_c \subset  \mathcal{V}$, corresponding to physical entities whose state changes causally govern the connectivity between spaces (e.g., doors, windows). At each time step $t$, the state $s_c^t$ of a control node $c \in V_c$ acts as a gating mechanism. The edge between an affected sensor pair $(i,j)$ is dynamically modulated to form the effective adjacency matrix $A^t$:
\begin{equation}
    A^t_{ij} = f_{\text{mod}}(s^t_c) \cdot A_{\text{Base}}(i,j)
\end{equation}
where $f_{\text{mod}}$ is a deterministic modulation function. For binary controls, we use an identity map ($f_{\text{mod}}(s)=s$), which translates the physical state into a topological switch that either permits or severs the interaction path. This localized, rule-based modulation allows for elegant handling of complex scenarios with multiple, independent control events.

This mechanism yields a sequence of adjacency matrices, ${A} = \{A^1, A^2, \dots, A^T\}$, where each $A^t$ offers a physically-grounded and context-aware snapshot of the network's true connectivity. This provides the downstream spatio-temporal layers with a far more realistic foundation for reasoning about dynamic events.

\subsection{Spatial Dependency Modeling}
Our spatial dependency module consists of $L$ stacked Graph Attention Network (GAT) layers designed to learn complex topological dependencies. To enable efficient computation, the input tensor $H^{(0)} \in {R}^{B \times T \times N \times D_{\text{Model}}}$ is reshaped to treat the time dimension as an extension of the batch, allowing $T$ independent graph snapshots to be processed in parallel.

Within each GAT layer, node features are updated by aggregating information from their neighbors, guided by our dynamically constructed graph $A^t$ as shown in Figure 4. The core of this process is a masked attention mechanism. First, attention coefficients $e_{ij,t}$ between nodes $i$ and $j$ are calculated using a shared linear transformation $W$ and a LeakyReLU-based scoring function. To ensure the GAT adheres strictly to the time-varying topology, we apply a mask derived from $A^t$. This mask sets the attention scores for all non-adjacent nodes to $-\infty$, effectively restricting information flow to only the edges present in the current graph snapshot.

These masked scores are then normalized row-wise using the Softmax function to produce the final attention weights $\alpha_{ij,t}$. The updated node representations for layer $l$ are computed via a weighted aggregation of the transformed neighbor features:
\begin{equation}
    H_t^{(l)} = \sigma\left(\sum_{j \in \mathcal{N}_i(t) \cup \{i\}} \alpha_{ij,t} \left(H_{j,t}^{(l-1)}W\right)\right)
\end{equation}
where $\mathcal{N}_i(t)$ is the set of neighbors of node $i$ at time $t$ according to $A^t$, and $\sigma$ is a non-linear activation function.

\begin{figure}[t]
    \centering
    \includegraphics[width=0.8\linewidth]{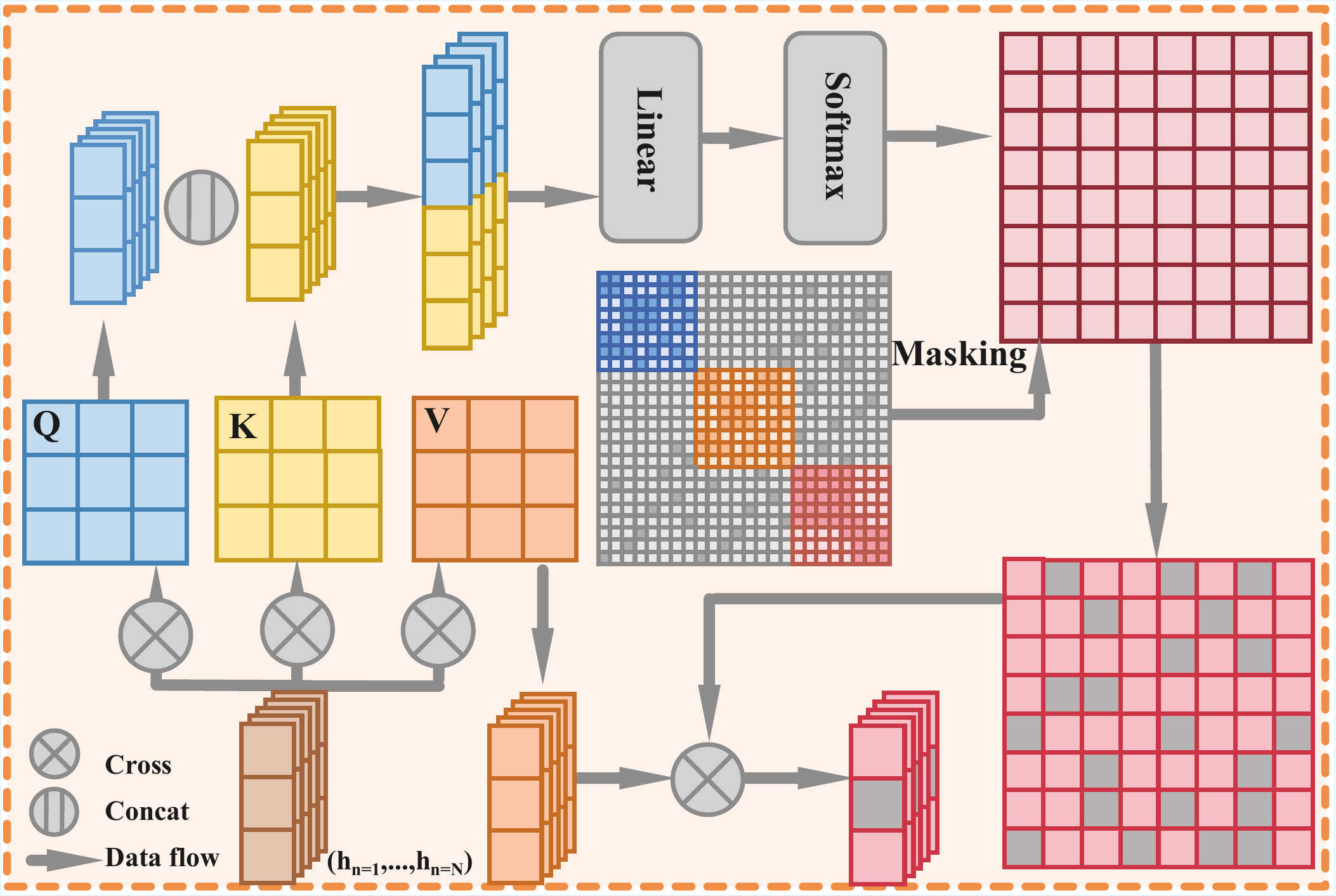}
    \caption{The dynamic Graph Attention mechanism. Attention weights are computed only over the active edges defined by the dynamic graph topology at a given time step.}
    \label{fig:gat_mechanism}
\end{figure}

After passing through $L$ such layers, the module outputs the spatially-aware feature tensor $H_{\text{spatial}}$. Each feature vector within this tensor is no longer an isolated sensor reading but a deep contextual summary, dynamically integrating information from its relevant multi-hop neighborhood at that specific moment.

\begin{figure*}[t]
    \centering
    \includegraphics[width=0.8\linewidth]{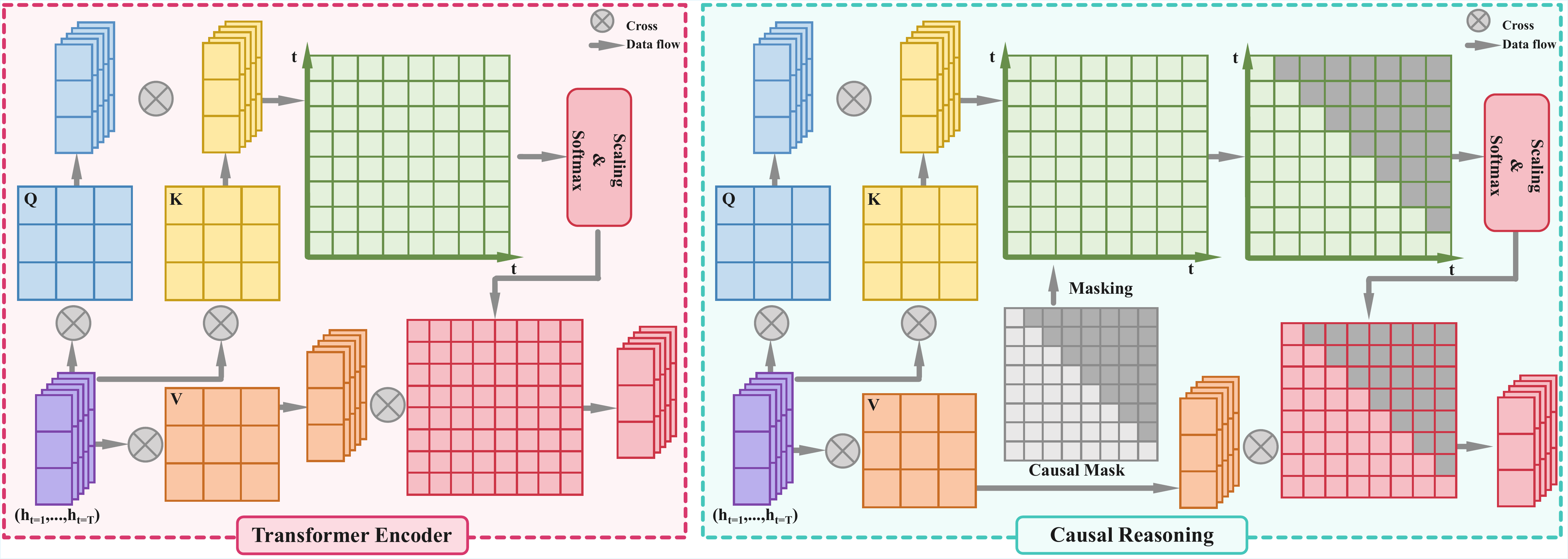}
    \caption{Comparison of information flow in the standard bidirectional Transformer (left) and Causal Reasoning module (right). The standard Transformer computes attention bidirectionally, capturing strong correlations between co-occurring events. In contrast, our Causal Reasoning module applies a causal mask, enforcing a unidirectional information flow, allowing the model to distinguish preceding events as more likely causes, thus moving beyond spurious correlations.}
    \label{fig5_introduction}
\end{figure*}

\subsection{Temporal Feature Extraction}
The spatially-aware feature tensor $H_{spatial}$ serves as the input to the temporal feature extraction module. The core of this module is architected as a multi-layer, multi-head bidirectional Transformer encoder. As shown in Figure 5, a key strategy is to logically treat the input tensor $H_{spatial}$ as a batch of $B \times N$ independent time series, each of length $T$. For each of these sequences, $(h_{(n,1)},h_{(n,2)},\dots,h_{(n,T)})$, the encoder computes bidirectional self-attention along the temporal dimension to capture its dynamics.\\
\indent Under this attention mechanism, each time step $t$ generates a Query vector, which is used to score its compatibility against the Key vectors of all other time steps within the sequence. This enables the model to construct a global attention map, allowing it to effectively capture long-range temporal dependencies and identify complex patterns.\\
\indent The module's final output is a deeply-fused feature tensor, which we term the Spatio-Temporal Representation, denoted as ${H}_{st}\in{R}^{B \times T \times N \times D_{\text{model}}}$. At this stage, each feature vector $h_{st,n,t}$ encapsulates the state of sensor n at time t, enriched with both its own holistic temporal context across the entire window and the comprehensive spatial context from its neighbors at all corresponding time steps.

\subsection{Causal Context Refinement}
Although the bidirectional representation $H_{st}$ is rich, it captures all correlations indiscriminately, including non-causal temporal co-occurrences. To address this, our Causal Context Refinement Module fundamentally redefines the model's temporal receptive field as shown in Figure 5.

Unlike a global, bidirectional field that accesses both past and future data, our approach imposes a strict causal structure. This is implemented via a multi-head self-attention mechanism governed by a causal mask. This mask strictly confines the receptive field for each time step $t$ to its historical context---that is, to information from steps $1, \dots, t$. This design enforces an autoregressive process, as formalized below:
\begin{equation}
    H_{\text{Causal}} = \text{MaskedSelfAttention}(H_{\text{st}})
\end{equation}
By constraining the information flow in this manner, the module forces the model to learn directional, cause-and-effect relationships rather than mere statistical associations. The resulting output, $H_{\text{Causal}} \in {R}^{B \times T \times N \times D_{\text{Model}}}$, is a representation firmly grounded in the arrow of time, making it inherently more robust against spurious correlations and providing a sounder basis for trustworthy predictions.

\subsection{Gated Fusion and Output Layer}
At this stage, we introduce a Gated Fusion Mechanism to intelligently integrate $H_{\text{st}}$ and $H_{\text{Causal}}$. A gate vector $G$ is generated by first concatenating both feature tensors and then passing them through a linear layer followed by a Sigmoid activation function:
\begin{equation}
    G = \sigma(W_g[H^{(L)} \Vert H_{\text{Causal}}] + b_g)
\end{equation}
where $\sigma$ is the Sigmoid function and $\Vert$ denotes the concatenation operation. Each element in the gate G is a value within the range (0, 1) that dynamically arbitrates between the two feature sources. The final fused representation, $H_{\text{fused}}$, is then computed as a weighted sum, modulated by the gate G:
\begin{equation}
    H_{\text{fused}} = G \odot H^{(L)} + (1 - G) \odot H_{\text{Causal}}
\end{equation}
Finally, $H_{\text{fused}}$ is passed through a two-layer feed-forward network, which serves as the prediction head, followed by a final Sigmoid function to produce the credibility score tensor, and the tensor is converted into a final classification by the thresholding mechanism.

\section{Experiment}
\subsection{Datasets}
This study utilizes two large-scale, real-world datasets, SHSD92 and SHSD104, collected within a dedicated smart home testbed. The experimental environment is equipped with 31 heterogeneous sensors—comprising 8 temperature, 8 humidity, 8 light, and 7 door sensors—which feature varying native sampling rates and are deployed according to the floor plan shown in Figure 2. Each dataset is collected over a one-month period, resulting in 2.5 GB of data in CSV format that captures a diverse range of human activity patterns such as cooking, bathing, and appliance use, with SHSD104 designed to present more complex scenarios.\\
\indent To establish a reliable ground truth, we employ a semi-automated process that combines manual annotation of observed hardware failures with the controlled injection of realistic anomalies (e.g., sensor drift and spikes). This process results in an overall anomaly ratio of approximately 15\%. Furthermore, more than 500 controlled dynamic events are precisely logged to provide the ground truth for validating our dynamic graph mechanism.
\subsection{Baselines}
We compare DyC-STG with 12 state-of-the-art (SOTA) baselines, carefully selected to represent the primary paradigms in spatio-temporal modeling. To ensure a comprehensive evaluation, we include: (1) Graph Neural Network-based models (DCRNN, STGCN, GWNET, MTGNN, STFGNN, STGNCDE \cite{li2017diffusion,yu2017spatio,li2021spatial,choi2022graph,tian2021spatial,wu2020connecting}), which are foundational for capturing structured dependencies; (2) Self-attention-based models (STTN, GMAN, ASTGCN, STJGCN \cite{ma2024spatial,zheng2020gman,guo2019attention,bai2020adaptive}), which leverage Transformer architectures; and (3) Dynamic Graph models (AGCRN, PDFormer \cite{zheng2023spatio,jiang2023pdformer}), which models evolving spatial correlations explicitly.

\subsection{Experiment settings}
\subsubsection{Dataset Processing.}
We first employ a Savitzky-Golay filter to smooth the raw sensor signals and mitigate high-frequency noise. All time series are then uniformly downsampled to 0.5 Hz for temporal alignment. We use a sliding-window approach to generate samples, where each sample consists of a 150-step (5-minute) historical context. The window slides in 15 steps. Crucially, our task is to leverage this 5-minute spatio-temporal context to assess the credibility of every data point within the same window, rather than forecasting future values. Finally, the generated samples are chronologically divided into training, validation, and testing sets with a ratio 70\% / 15\% / 15\%.
\subsubsection{Model Settings.}
All models were trained for 100 epochs on an NVIDIA RTX 4060 GPU (16GB) using PyTorch 2.2.1. We employed the AdamW optimizer with an initial learning rate of $0.001$, a weight decay of $0.0001$, a batch size of 32, and a Cosine Annealing scheduler for dynamic learning rate adjustment.

To address the inherent class imbalance, we utilized a Focal Loss function with parameters $\alpha=0.75$ and $\gamma=2.0$. For our DyC-STG architecture, the optimal hyperparameters ($d_{model}=128$, 4 attention heads) were determined through a grid search evaluated on the validation set. Baseline models adopted the hyperparameter settings from their original publications, with only their output layers adapted for our classification task.

\subsubsection{Evaluation Metrics.}
We assess model performance using four standard classification metrics: Precision, Recall, F1-Score, and AUC (Area Under the Receiver Operating Characteristic Curve). The F1-Score offers a balanced measure between Precision and Recall. AUC provides a comprehensive evaluation of the model's ability to distinguish between classes across all possible thresholds.

\subsection{Performance Comparison}
The overall performance of our proposed DyC-STG model against all baselines on the SHSD92 and SHSD104 datasets is presented in Table 1. The results unequivocally demonstrate the superiority of our framework across all evaluation metrics on both datasets.
\begin{table*}[t]
    \centering 
    \begin{threeparttable}
    
    \begin{tabular}{llcccccccc}
        \toprule
        \multirow{2}{*}{Model} & \multirow{2}{*}{Model Variant} & \multicolumn{4}{c}{SHSD92} & \multicolumn{4}{c}{SHSD104} \\
        \cmidrule(lr){3-6} \cmidrule(lr){7-10} 
         & & Precision & Recall & F1-Score & AUC & Precision & Recall & F1-Score & AUC \\
        \midrule
        
        DCRNN& DConv+GRU & 0.8615 & 0.8321 & 0.8466 & 0.9588 & 0.8251 & 0.7835 & 0.8038 & 0.9255 \\
        STGCN& GCN+TCN & 0.8559 & 0.8294 & 0.8424 & 0.9543 & 0.8176 & 0.7769 & 0.7967 & 0.9212 \\
        GWNet& Graph+TCN & 0.9152 & 0.8718 & 0.8929 & 0.9785 & 0.8813 & 0.8351 & 0.8576 & 0.9593 \\
        MTGNN& Graph+GCN & 0.9031 & 0.8665 & 0.8844 & 0.9743 & 0.8655 & 0.8288 & 0.8467 & 0.9519 \\
        STFGNN& TFusion+GNN & 0.8995 & 0.8501 & 0.8741 & 0.9702 & 0.8596 & 0.8105 & 0.8343 & 0.9471 \\
        STGNCDE& GCN+CDE & 0.9088 & 0.8473 & 0.8770 & 0.9731 & 0.8724 & 0.8152 & 0.8428 & 0.9524 \\
        STTN& ST-Trans & 0.9105 & 0.8699 & 0.8897 & 0.9769 & 0.8858 & 0.8395 & 0.8620 & 0.9601 \\
        GMAN& ST-Attn & 0.9026 & 0.8753 & 0.8887 & 0.9754 & 0.8715 & 0.8411 & 0.8560 & 0.9567 \\
        ASTGCN& Attn+GCN+GRU & 0.8749 & 0.8402 & 0.8572 & 0.9634 & 0.8392 & 0.7955 & 0.8167 & 0.9318 \\
        AGCRN& GCN+GRU & \textbf{0.9599} & 0.8217 & 0.8856 & \underline{0.9835} & \underline{0.9387} & 0.8021 & 0.8651 & \underline{0.9712} \\
        STJGCJ& Graph+Attn & 0.8813 & 0.8564 & 0.8687 & 0.9677 & 0.8488 & 0.8117 & 0.8298 & 0.9396 \\
        PDFormer& Graph+Trans & 0.9288 & \textbf{0.9021} & \underline{0.9153} & 0.9819 & 0.9075 & \textbf{0.8996} & \underline{0.9035} & 0.9684 \\
        \rowcolor{gray!20}
        DyC-STG & GAT+Trans & \underline{0.9597} & \underline{0.9015} & \textbf{0.9297} & \textbf{0.9886} & \textbf{0.9485} & \underline{0.8912} & \textbf{0.9189} & \textbf{0.9823} \\
        \bottomrule
    \end{tabular}

    \caption{Quantitative comparison of DyC-STG against state-of-the-art baselines on the data credibility assessment task. \textbf{Bold} and \underline{underlined} values indicate the \textbf{best} and \underline{second-best} results, respectively.}
    \label{tab:performance}


    \end{threeparttable}
\end{table*}
\begin{table*}[t]
    \centering 
    
    \setlength{\tabcolsep}{3pt}
    \begin{tabular}{cccccccccccc}
        \toprule
\multirow{2}{1.3cm}{\centering Dynamic\\Graph} & 
\multirow{2}{0.9cm}{\centering GAT} & 
\multirow{2}{1.7cm}{\centering Transformer\\encoder} & 
\multirow{2}{1.7cm}{\centering Causal\\Reasoning} & 
\multicolumn{4}{c}{SHSD92} & 
\multicolumn{4}{c}{SHSD104} \\

\cmidrule(lr){5-8} \cmidrule(lr){9-12}
        
         & & & & 
        Precision & Recall & F1-Score & AUC & 
        Precision & Recall & F1-Score & AUC \\
        \midrule
        
\rowcolor{gray!20}
$\checkmark$ & $\checkmark$ & $\checkmark$ & $\checkmark$ & \textbf{0.9597} & \textbf{0.9015} & \textbf{0.9297} & \textbf{0.9886} & 0.9485 & \textbf{0.8912} & \textbf{0.9189} & \textbf{0.9823} \\
$\times$     & $\checkmark$ & $\checkmark$ & $\checkmark$ & 0.9046 & 0.8585 & 0.8810 & 0.9717 & 0.9088 & 0.8415 & 0.8739 & 0.9665 \\
$\checkmark$ & $\times$     & $\checkmark$ & $\checkmark$ & 0.9557 & 0.8520 & 0.9009 & 0.9769 & \textbf{0.9521} & 0.8577 & 0.9024 & 0.9748 \\
$\checkmark$ & $\checkmark$ & $\times$     & $\checkmark$ & 0.9373 & 0.7721 & 0.8467 & 0.9559 & 0.9405 & 0.7823 & 0.8546 & 0.9585 \\
$\checkmark$ & $\checkmark$ & $\checkmark$ & $\times$     & 0.8583 & 0.7955 & 0.8257 & 0.9534 & 0.8421 & 0.8066 & 0.8239 & 0.9521 \\
        \bottomrule
    \end{tabular}
    
    \caption{Ablation study of model components on the SHSD92 and SHSD104 datasets. \textbf{Bold} value indicates the \textbf{best} result.}
    \label{tab:model_components}

\end{table*}
The results in Table~\ref{tab:performance} demonstrate that DyC-STG significantly outperforms all baselines, establishing a new state-of-the-art with an F1-Score of 0.9297 and an AUC of 0.9886 on SHSD92. This success is part of a clear trend where dynamic graph models (including PDFormer, AGCRN) substantially outperform static methods, validating our core hypothesis on the necessity of modeling dynamic dependencies. Furthermore, DyC-STG exhibits superior robustness on the more challenging SHSD104 dataset, where its F1-Score drops by a mere 1.2\% compared to a 5.4\% decline for classic models like STGCN, highlighting its strong generalization capabilities.

DyC-STG's advantage, particularly over other dynamic models like PDFormer, stems from its unique architecture. The event-driven dynamic graph provides a more principled model of physical state changes, while the causal reasoning module disambiguates true cause-and-effect from mere correlation. This combination enables a more fundamental and robust understanding of the underlying system dynamics.

\subsection{Ablation Study}
\subsubsection{Analysis of Model Components.}
To validate the contribution of each component within DyC-STG, we conducted a comprehensive ablation study, with the results presented in Table~\ref{tab:model_components}. The findings confirm that all modules are integral to the model's success, as the removal of any single one leads to a significant degradation in performance. The key insights are as follows:

\noindent \textbf{Causal Reasoning:} This module proves to be the most critical component. Its removal incurs the most substantial performance drop, with the F1-Score on SHSD92 plummeting by 10.4\%. This starkly demonstrates the necessity of refining spatio-temporal features to enforce temporal causality and filter out spurious correlations.

\noindent \textbf{Dynamic Graph:} The dynamic graph mechanism is essential. Replacing our event-driven dynamic graph with a static one results in a significant 4.87\% decrease in F1-Score, which quantitatively validates our core hypothesis on the importance of adapting the graph structure to physical events.

\noindent \textbf{Transformer Encoder:} The Transformer is vital for modeling temporal context. Its removal causes an 8.3\% drop in F1-Score, highlighting the importance of the architecture for capturing complex, long-range temporal dependencies.
    
\noindent \textbf{Graph Attention (GAT):} Finally, omitting the GAT layers also diminishes performance. This validates our choice of an anisotropic spatial aggregation mechanism, which allows the model to learn different importance weights for various neighbors.

\begin{figure}[t]
    \centering
    \includegraphics[width=1.0\linewidth]{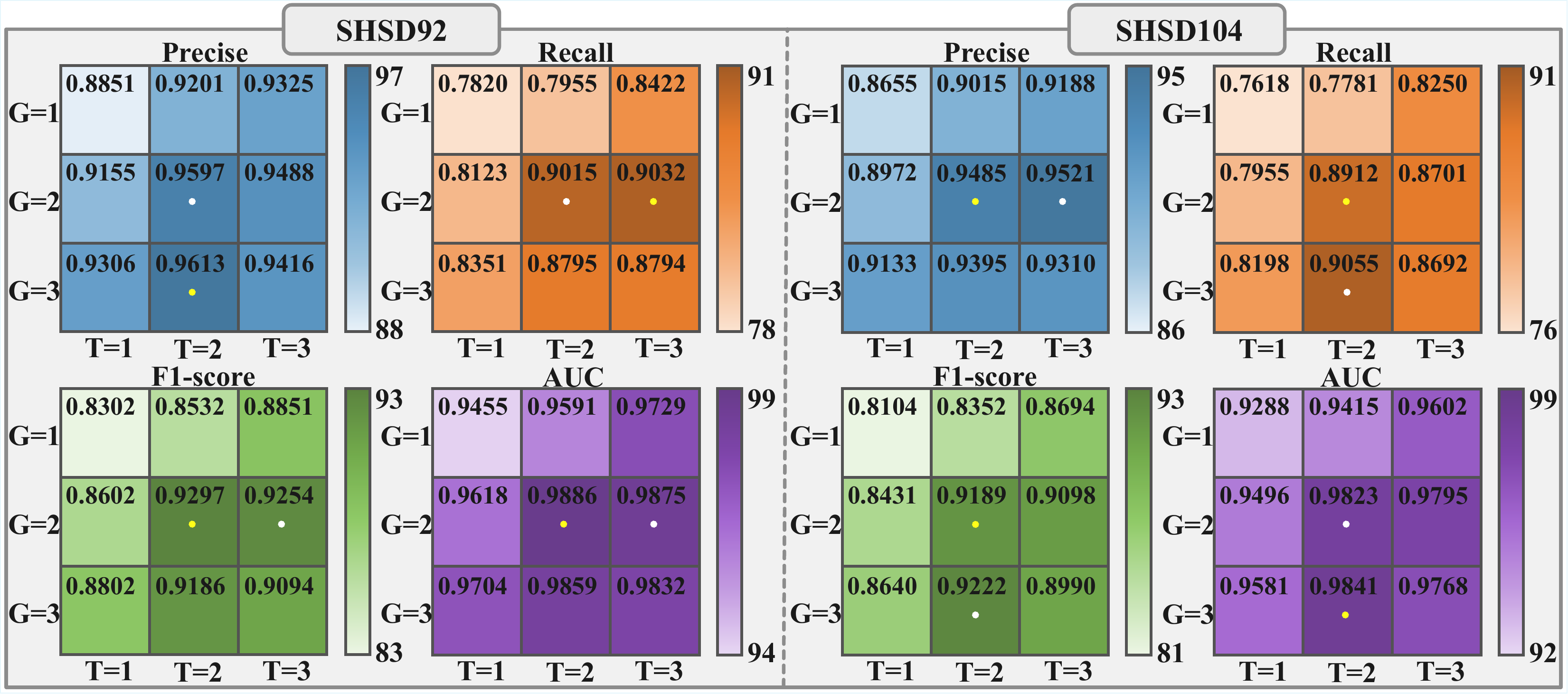}
    \caption{Evaluation results of the impact of the number of model layers of GAT and Transformer encoder. The yellow/white spot indicates the best/second-best result.
}
    \label{fig6_introduction}
\end{figure} 

\subsubsection{Number of DyC-STG Layers.}
Figure 6 visualizes the impact of varying the number of GAT layers ($G_l$) and Transformer encoder layers ($T_l$) on model performance.  The results across both datasets consistently reveal a non-linear relationship between model depth and efficacy.  A distinct "sweet spot" emerges at a moderate depth, with the ($G_l$=2, $T_l$=2) configuration achieving the optimal F1-Score of 0.9297 on the SHSD92 dataset.  Shallower architectures (i.e., $G_l$=1 or $T_l$=1) suffer from significant performance degradation, likely due to insufficient capacity to capture complex spatio-temporal dependencies.  Conversely, increasing the depth beyond two layers (e.g., to $G_l$=3, $T_l$=3) leads to diminishing returns or even a slight performance decline.  This is likely attributable to challenges such as over-smoothing in deeper GAT layers and increased optimization difficulty in deeper Transformers.  This analysis empirically validates our choice of a 2+2 layer architecture, as it strikes an optimal balance between model capacity and performance, avoiding both underfitting and the adverse effects of excessive depth.

\section{Conclusion}
In this paper, we introduce DyC-STG, a novel framework designed to overcome the limitations of existing models in assessing IoT data credibility. By integrating an event-driven dynamic graph with a causality-enhanced Transformer, DyC-STG adapts to physical state changes while distinguishing causal drivers from spurious correlations. DyC-STG achieves state-of-the-art performance on two real-world smart home datasets. Ablation studies empirically validate our central hypothesis: both the dynamic graph adaptation and the causal reasoning module are critical to its superior performance. Future work will focus on enhancing the framework’s autonomy by automatically learning event-driven graph dynamics, and extending this physically-grounded paradigm to other domains like industrial IoT.

\bibliography{aaai2026}

\begin{thebibliography}{43}
\providecommand{\natexlab}[1]{#1}

\bibitem[{Aky{\"u}rek et~al.(2024)Aky{\"u}rek, Wang, Kim, and Andreas}]{akyurek2024context}
Aky{\"u}rek, E.; Wang, B.; Kim, Y.; and Andreas, J. 2024.
\newblock In-context language learning: architectures and algorithms.
\newblock In \emph{Proceedings of the 41st International Conference on Machine Learning}, 787--812.

\bibitem[{Aouedi et~al.(2024)Aouedi, Vu, Sacco, Nguyen, Piamrat, Marchetto, and Pham}]{aouedi2024survey}
Aouedi, O.; Vu, T.-H.; Sacco, A.; Nguyen, D.~C.; Piamrat, K.; Marchetto, G.; and Pham, Q.-V. 2024.
\newblock A survey on intelligent Internet of Things: Applications, security, privacy, and future directions.
\newblock \emph{IEEE communications surveys \& tutorials}.

\bibitem[{Bai et~al.(2020)Bai, Yao, Li, Wang, and Wang}]{bai2020adaptive}
Bai, L.; Yao, L.; Li, C.; Wang, X.; and Wang, C. 2020.
\newblock Adaptive graph convolutional recurrent network for traffic forecasting.
\newblock \emph{Advances in neural information processing systems}, 33: 17804--17815.

\bibitem[{Chen et~al.(2023)Chen, Chen, Shang, Wu, Zheng, Wen, and Zhang}]{chen2023multi}
Chen, L.; Chen, D.; Shang, Z.; Wu, B.; Zheng, C.; Wen, B.; and Zhang, W. 2023.
\newblock Multi-scale adaptive graph neural network for multivariate time series forecasting.
\newblock \emph{IEEE Transactions on Knowledge and Data Engineering}, 35(10): 10748--10761.

\bibitem[{Chen et~al.(2024)Chen, Sun, Saluz, Schiavon, and Geyer}]{chen2024using}
Chen, X.; Sun, R.; Saluz, U.; Schiavon, S.; and Geyer, P. 2024.
\newblock Using causal inference to avoid fallouts in data-driven parametric analysis: A case study in the architecture, engineering, and construction industry.
\newblock \emph{Developments in the Built Environment}, 17: 100296.

\bibitem[{Choi et~al.(2022)Choi, Choi, Hwang, and Park}]{choi2022graph}
Choi, J.; Choi, H.; Hwang, J.; and Park, N. 2022.
\newblock Graph neural controlled differential equations for traffic forecasting.
\newblock In \emph{Proceedings of the AAAI conference on artificial intelligence}, volume~36, 6367--6374.

\bibitem[{Fu, Pan, and Zhang(2024)}]{fu2024causal}
Fu, X.; Pan, Y.; and Zhang, L. 2024.
\newblock A causal-temporal graphic convolutional network (CT-GCN) approach for TBM load prediction in tunnel excavation.
\newblock \emph{Expert Systems with Applications}, 238: 121977.

\bibitem[{Geng et~al.(2024)Geng, Xu, Wu, Zhao, Wang, Li, and Zhang}]{geng2024stgaformer}
Geng, Z.; Xu, J.; Wu, R.; Zhao, C.; Wang, J.; Li, Y.; and Zhang, C. 2024.
\newblock STGAFormer: Spatial--temporal gated attention transformer based graph neural network for traffic flow forecasting.
\newblock \emph{Information Fusion}, 105: 102228.

\bibitem[{Gong et~al.(2024)Gong, Zhang, Yao, Bi, Li, and Xu}]{gong2024causal}
Gong, C.; Zhang, C.; Yao, D.; Bi, J.; Li, W.; and Xu, Y. 2024.
\newblock Causal discovery from temporal data: An overview and new perspectives.
\newblock \emph{ACM Computing Surveys}, 57(4): 1--38.

\bibitem[{Guo et~al.(2019)Guo, Lin, Feng, Song, and Wan}]{guo2019attention}
Guo, S.; Lin, Y.; Feng, N.; Song, C.; and Wan, H. 2019.
\newblock Attention based spatial-temporal graph convolutional networks for traffic flow forecasting.
\newblock In \emph{Proceedings of the AAAI conference on artificial intelligence}, volume~33, 922--929.

\bibitem[{Guo et~al.(2024)Guo, Chen, Wang, Chang, Pei, Chawla, Wiest, and Zhang}]{guo2024large}
Guo, T.; Chen, X.; Wang, Y.; Chang, R.; Pei, S.; Chawla, N.~V.; Wiest, O.; and Zhang, X. 2024.
\newblock Large Language Model Based Multi-agents: A Survey of Progress and Challenges.
\newblock In \emph{IJCAI}.

\bibitem[{Huda et~al.(2024)Huda, Ahmed, Adnan, Ali, and Naeem}]{huda2024experts}
Huda, N.~U.; Ahmed, I.; Adnan, M.; Ali, M.; and Naeem, F. 2024.
\newblock Experts and intelligent systems for smart homes’ Transformation to Sustainable Smart Cities: A comprehensive review.
\newblock \emph{Expert Systems with Applications}, 238: 122380.

\bibitem[{Jiang et~al.(2023)Jiang, Han, Zhao, and Wang}]{jiang2023pdformer}
Jiang, J.; Han, C.; Zhao, W.~X.; and Wang, J. 2023.
\newblock Pdformer: Propagation delay-aware dynamic long-range transformer for traffic flow prediction.
\newblock In \emph{Proceedings of the AAAI conference on artificial intelligence}, volume~37, 4365--4373.

\bibitem[{Jin et~al.(2023)Jin, Liang, Fang, Shao, Huang, Zhang, and Zheng}]{jin2023spatio}
Jin, G.; Liang, Y.; Fang, Y.; Shao, Z.; Huang, J.; Zhang, J.; and Zheng, Y. 2023.
\newblock Spatio-temporal graph neural networks for predictive learning in urban computing: A survey.
\newblock \emph{IEEE transactions on knowledge and data engineering}, 36(10): 5388--5408.

\bibitem[{Jing et~al.(2024)Jing, Zhou, Ren, and Yang}]{jing2024causality}
Jing, B.; Zhou, D.; Ren, K.; and Yang, C. 2024.
\newblock Causality-aware spatiotemporal graph neural networks for spatiotemporal time series imputation.
\newblock In \emph{Proceedings of the 33rd ACM International Conference on Information and Knowledge Management}, 1027--1037.

\bibitem[{Ju et~al.(2024)Ju, Zhao, Qin, Yi, Yuan, Xiao, Luo, Yan, and Zhang}]{ju2024cool}
Ju, W.; Zhao, Y.; Qin, Y.; Yi, S.; Yuan, J.; Xiao, Z.; Luo, X.; Yan, X.; and Zhang, M. 2024.
\newblock Cool: a conjoint perspective on spatio-temporal graph neural network for traffic forecasting.
\newblock \emph{Information Fusion}, 107: 102341.

\bibitem[{Kapoor et~al.(2024)Kapoor, Schulz, Vetter, Pei, Gao, and Macke}]{kapoor2024latent}
Kapoor, J.; Schulz, A.; Vetter, J.; Pei, F.; Gao, R.; and Macke, J.~H. 2024.
\newblock Latent diffusion for neural spiking data.
\newblock \emph{Advances in Neural Information Processing Systems}, 37: 118119--118154.

\bibitem[{Kong, Guo, and Liu(2024)}]{kong2024spatio}
Kong, W.; Guo, Z.; and Liu, Y. 2024.
\newblock Spatio-temporal pivotal graph neural networks for traffic flow forecasting.
\newblock In \emph{Proceedings of the AAAI conference on artificial intelligence}, volume~38, 8627--8635.

\bibitem[{Li and Zhu(2021)}]{li2021spatial}
Li, M.; and Zhu, Z. 2021.
\newblock Spatial-temporal fusion graph neural networks for traffic flow forecasting.
\newblock In \emph{Proceedings of the AAAI conference on artificial intelligence}, volume~35, 4189--4196.

\bibitem[{Li et~al.(2017)Li, Yu, Shahabi, and Liu}]{li2017diffusion}
Li, Y.; Yu, R.; Shahabi, C.; and Liu, Y. 2017.
\newblock Diffusion convolutional recurrent neural network: Data-driven traffic forecasting.
\newblock \emph{arXiv preprint arXiv:1707.01926}.

\bibitem[{Li et~al.(2025)Li, Hu, Han, Gu, and Cai}]{li2025ssl}
Li, Z.; Hu, Z.; Han, P.; Gu, Y.; and Cai, S. 2025.
\newblock SSL-STMFormer Self-Supervised Learning Spatio-Temporal Entanglement Transformer for Traffic Flow Prediction.
\newblock In \emph{Proceedings of the AAAI Conference on Artificial Intelligence}, volume~39, 12130--12138.

\bibitem[{Liu and Zhang(2024)}]{liu2024spatial}
Liu, A.; and Zhang, Y. 2024.
\newblock Spatial--temporal dynamic graph convolutional network with interactive learning for traffic forecasting.
\newblock \emph{IEEE Transactions on Intelligent Transportation Systems}, 25(7): 7645--7660.

\bibitem[{Ma, Zhao, and Hou(2024)}]{ma2024spatial}
Ma, J.; Zhao, J.; and Hou, Y. 2024.
\newblock Spatial--temporal transformer networks for traffic flow forecasting using a pre-trained language model.
\newblock \emph{Sensors}, 24(17): 5502.

\bibitem[{Nichani, Damian, and Lee(2024)}]{nichani2024transformers}
Nichani, E.; Damian, A.; and Lee, J.~D. 2024.
\newblock How transformers learn causal structure with gradient descent.
\newblock In \emph{Proceedings of the 41st International Conference on Machine Learning}, 38018--38070.

\bibitem[{Nikpour et~al.(2025)Nikpour, Yousefi, Jafarzadeh, Danesh, Shomali, Asadi, Lonbar, and Ahmadi}]{nikpour2025intelligent}
Nikpour, M.; Yousefi, P.~B.; Jafarzadeh, H.; Danesh, K.; Shomali, R.; Asadi, S.; Lonbar, A.~G.; and Ahmadi, M. 2025.
\newblock Intelligent energy management with iot framework in smart cities using intelligent analysis: An application of machine learning methods for complex networks and systems.
\newblock \emph{Journal of Network and Computer Applications}, 235: 104089.

\bibitem[{Paramesha, Rane, and Rane(2024)}]{paramesha2024big}
Paramesha, M.; Rane, N.; and Rane, J. 2024.
\newblock Big data analytics, artificial intelligence, machine learning, internet of things, and blockchain for enhanced business intelligence.
\newblock \emph{Artificial Intelligence, Machine Learning, Internet of Things, and Blockchain for Enhanced Business Intelligence (June 6, 2024)}.

\bibitem[{Pazho et~al.(2023)Pazho, Noghre, Purkayastha, Vempati, Martin, and Tabkhi}]{pazho2023survey}
Pazho, A.~D.; Noghre, G.~A.; Purkayastha, A.~A.; Vempati, J.; Martin, O.; and Tabkhi, H. 2023.
\newblock A survey of graph-based deep learning for anomaly detection in distributed systems.
\newblock \emph{IEEE Transactions on Knowledge and Data Engineering}, 36(1): 1--20.

\bibitem[{Qiu et~al.(2024)Qiu, Xie, Ji, Liu, and Wang}]{qiu2024integrating}
Qiu, Z.; Xie, Z.; Ji, Z.; Liu, X.; and Wang, G. 2024.
\newblock Integrating query data for enhanced traffic forecasting: A Spatio-Temporal Graph Attention Convolution Network approach with delay modeling.
\newblock \emph{Knowledge-Based Systems}, 301: 112315.

\bibitem[{Rehman et~al.(2024)Rehman, Gondal, Ge, Dong, Gregory, and Tari}]{rehman2024proactive}
Rehman, Z.; Gondal, I.; Ge, M.; Dong, H.; Gregory, M.; and Tari, Z. 2024.
\newblock Proactive defense mechanism: Enhancing IoT security through diversity-based moving target defense and cyber deception.
\newblock \emph{Computers \& Security}, 139: 103685.

\bibitem[{Shao et~al.(2025)Shao, Wang, Yao, Bell, and Gao}]{shao2025st}
Shao, Z.; Wang, Z.; Yao, X.; Bell, M.~G.; and Gao, J. 2025.
\newblock ST-MambaSync: Complement the power of Mamba and Transformer fusion for less computational cost in spatial--temporal traffic forecasting.
\newblock \emph{Information Fusion}, 117: 102872.

\bibitem[{Sun et~al.(2025)Sun, Hu, Gu, Chen, Liang, and Yang}]{sun2025scalable}
Sun, C.; Hu, J.; Gu, H.; Chen, J.; Liang, W.; and Yang, M. 2025.
\newblock Scalable and adaptive graph neural networks with self-label-enhanced training.
\newblock \emph{Pattern Recognition}, 160: 111210.

\bibitem[{Tang et~al.(2023)Tang, Hua, Gao, Zhao, and Li}]{tang2023gadbench}
Tang, J.; Hua, F.; Gao, Z.; Zhao, P.; and Li, J. 2023.
\newblock Gadbench: Revisiting and benchmarking supervised graph anomaly detection.
\newblock \emph{Advances in Neural Information Processing Systems}, 36: 29628--29653.

\bibitem[{Tian and Chan(2021)}]{tian2021spatial}
Tian, C.; and Chan, W.~K. 2021.
\newblock Spatial-temporal attention wavenet: A deep learning framework for traffic prediction considering spatial-temporal dependencies.
\newblock \emph{IET Intelligent Transport Systems}, 15(4): 549--561.

\bibitem[{Verma et~al.(2024)Verma, Boonsanong, Hoang, Hines, Dickerson, and Shah}]{verma2024counterfactual}
Verma, S.; Boonsanong, V.; Hoang, M.; Hines, K.; Dickerson, J.; and Shah, C. 2024.
\newblock Counterfactual explanations and algorithmic recourses for machine learning: A review.
\newblock \emph{ACM Computing Surveys}, 56(12): 1--42.

\bibitem[{Wang et~al.(2024{\natexlab{a}})Wang, Xin, Zhang, Perez-Cruz, and Raubal}]{wang2024counterfactual}
Wang, R.; Xin, Y.; Zhang, Y.; Perez-Cruz, F.; and Raubal, M. 2024{\natexlab{a}}.
\newblock Counterfactual explanations for deep learning-based traffic forecasting.
\newblock \emph{arXiv preprint arXiv:2405.00456}.

\bibitem[{Wang et~al.(2024{\natexlab{b}})Wang, Wang, Jia, Zhang, Klimenko, Wang, He, Huang, and Liu}]{wang2024spatiotemporal}
Wang, Z.; Wang, Y.; Jia, F.; Zhang, F.; Klimenko, N.; Wang, L.; He, Z.; Huang, Z.; and Liu, Y. 2024{\natexlab{b}}.
\newblock Spatiotemporal Fusion Transformer for large-scale traffic forecasting.
\newblock \emph{Information Fusion}, 107: 102293.

\bibitem[{Wu et~al.(2020)Wu, Pan, Long, Jiang, Chang, and Zhang}]{wu2020connecting}
Wu, Z.; Pan, S.; Long, G.; Jiang, J.; Chang, X.; and Zhang, C. 2020.
\newblock Connecting the dots: Multivariate time series forecasting with graph neural networks.
\newblock In \emph{Proceedings of the 26th ACM SIGKDD international conference on knowledge discovery \& data mining}, 753--763.

\bibitem[{Xiao et~al.(2023)Xiao, Zou, Li, Zhao, Li, Weng, Li, and Jiang}]{xiao2023know}
Xiao, J.; Zou, Q.; Li, Q.; Zhao, D.; Li, K.; Weng, Z.; Li, R.; and Jiang, Y. 2023.
\newblock I know your intent: Graph-enhanced intent-aware user device interaction prediction via contrastive learning.
\newblock \emph{Proceedings of the ACM on Interactive, Mobile, Wearable and Ubiquitous Technologies}, 7(3): 1--28.

\bibitem[{Xie et~al.(2023)Xie, Ma, Li, Ji, Du, Yu, and Zhang}]{xie2023spatio}
Xie, P.; Ma, M.; Li, T.; Ji, S.; Du, S.; Yu, Z.; and Zhang, J. 2023.
\newblock Spatio-temporal dynamic graph relation learning for urban metro flow prediction.
\newblock \emph{IEEE Transactions on Knowledge and Data Engineering}, 35(10): 9973--9984.

\bibitem[{Yu, Yin, and Zhu(2017)}]{yu2017spatio}
Yu, B.; Yin, H.; and Zhu, Z. 2017.
\newblock Spatio-temporal graph convolutional networks: A deep learning framework for traffic forecasting.
\newblock \emph{arXiv preprint arXiv:1709.04875}.

\bibitem[{Zhang and Wu(2024)}]{zhang2024graph}
Zhang, Z.; and Wu, L. 2024.
\newblock Graph neural network-based bearing fault diagnosis using Granger causality test.
\newblock \emph{Expert Systems with Applications}, 242: 122827.

\bibitem[{Zheng et~al.(2023)Zheng, Fan, Pan, Jin, Peng, Wu, Wang, and Yu}]{zheng2023spatio}
Zheng, C.; Fan, X.; Pan, S.; Jin, H.; Peng, Z.; Wu, Z.; Wang, C.; and Yu, P.~S. 2023.
\newblock Spatio-temporal joint graph convolutional networks for traffic forecasting.
\newblock \emph{IEEE Transactions on Knowledge and Data Engineering}, 36(1): 372--385.

\bibitem[{Zheng et~al.(2020)Zheng, Fan, Wang, and Qi}]{zheng2020gman}
Zheng, C.; Fan, X.; Wang, C.; and Qi, J. 2020.
\newblock Gman: A graph multi-attention network for traffic prediction.
\newblock In \emph{Proceedings of the AAAI conference on artificial intelligence}, volume~34, 1234--1241.

\end{thebibliography}
\input{}

\end{document}